\documentclass{article}

\PassOptionsToPackage{numbers, compress}{natbib}
\usepackage[final]{neurips_2025}

\usepackage[utf8]{inputenc} 
\usepackage[T1]{fontenc}    
\usepackage[hidelinks]{hyperref}       
\usepackage{url}            
\usepackage{booktabs}       
\usepackage{amsfonts}       
\usepackage{nicefrac}       
\usepackage{microtype}      
\usepackage{xcolor}         
\usepackage{multirow}  
\usepackage{threeparttable}
\usepackage{xspace}
\usepackage{adjustbox} 
\usepackage{subcaption}

\usepackage{amsmath}
\usepackage{cleveref}
\usepackage{graphicx}
\usepackage{enumitem}
\usepackage{marvosym}
\usepackage{makecell}
\usepackage{arydshln}

\usepackage{pifont}
\newcommand{\xmark}{\ding{55}}
\newcommand{\coloredcheckmark}{\checkmark}

\newcommand{\dipro}{\texttt{DiPro}\xspace}

\title{Multimodal Disease Progression Modeling via Spatio-\\temporal Disentanglement and Multiscale Alignment}

\author{
    Chen Liu\textsuperscript{ \rm 1}, \
    Wenfang Yao\textsuperscript{\rm 1},\ 
    Kejing Yin\textsuperscript{\rm 2}\thanks{Correspondence to: Kejing Yin <cskjyin@comp.hkbu.edu.hk>}, \ 
    William K. Cheung\textsuperscript{\rm 2},\ 
    Jing Qin\textsuperscript{\rm 1}\\
  \textsuperscript{\rm 1}School of Nursing, The Hong Kong Polytechnic University\\
  \textsuperscript{\rm 2}Department of Computer Science, Hong Kong Baptist University
}

\begin{document}

\maketitle

\begin{abstract}
Longitudinal multimodal data, including electronic health records (EHR) and sequential chest X-rays (CXRs), is critical for modeling disease progression, yet remains underutilized due to two key challenges: (1) redundancy in consecutive CXR sequences, where static anatomical regions dominate over clinically-meaningful dynamics, and (2) temporal misalignment between sparse, irregular imaging and continuous EHR data. We introduce \dipro, a novel framework that addresses these challenges through region-aware disentanglement and multi-timescale alignment. First, we disentangle static (anatomy) and dynamic (pathology progression) features in sequential CXRs, prioritizing disease-relevant changes. Second, we hierarchically align these static and dynamic CXR features with asynchronous EHR data via local (pairwise interval-level) and global (full-sequence) synchronization to model coherent progression pathways. Extensive experiments on the MIMIC dataset demonstrate that \dipro could effectively extract temporal clinical dynamics and achieve state-of-the-art performance on both disease progression identification and general ICU prediction tasks.

\end{abstract}

\section{Introduction}

Accurately modeling disease progression, i.e., the temporal evolution of a disease, is critical for personalized clinical care decision-making~\cite{mould2022disease, moor2021early, zhong2017perspective}. By capturing progression dynamics, predictive models can enable early identification of deterioration, improve precise risk stratification, and inform individualized treatment planning~\cite{moor2021early, ghazi2019training, wang2014unsupervised, konerman2019machine}.  In ICU settings, for instance, tracking sepsis progression through multimodal data (e.g., vitals, labs, and organ functions) is essential for identifying early deterioration and initiating lifesaving treatments~\cite{seymour2017time, singer2016third}. 

Modern healthcare increasingly relies on longitudinal clinical data to track disease progression. Sequential chest X-rays (CXRs) capture valuable visual evidence of anatomical and pathological changes over time, while electronic health records (EHRs) provide continuous physiological metrics and treatment responses~\cite{swinckels2024use, cascarano2023machine}. The complementary nature of these modalities offers a unique opportunity: a fusion of imaging and clinical time-series data could enable a more holistic modeling of disease trajectories~\cite{harutyunyan2019multitask, zhou2024multimodal}.
Despite a growing number of studies that explored disease progression modeling and multimodal fusion using longitudinal clinical data, two key challenges remain insufficiently addressed:
\paragraph{Redundancy in clinical image sequences.}
Static anatomical features (e.g., chronic cardiac enlargement or stable skeletal deformities) dominate sequential CXR scans, often obscuring subtle but clinically important pathological changes (e.g., new infiltrates or evolving edema). Without explicit disentanglement, the signals of disease progression become diluted, reducing the utility of sequential imaging analysis. 
Recent CXR-based progression models (e.g.,~\cite{zhu2024symptom, karwande2022chexrelnet}) primarily focus on extracting symptom-related features and anatomical-disease co-occurrences, treating all imaging features uniformly. However, distinguishing long-term anatomical baselines from evolving pathological changes is the key to improving the precision of disease progression modeling, yet is largely overlooked~\cite{dicks2019modeling, mcevoy2011mild}. 
\paragraph{Temporal misalignments across modalities.} While EHRs provide continuous, high-frequency measurements (e.g., hourly vitals or lab tests), CXRs offer only sparse, irregular snapshots, creating an intrinsic misalignment in timescales. This discrepancy in temporal granularity complicates the alignment of cross-modal trends and may obscure short-term clinical deteriorations between imaging intervals. Existing multimodal fusion approaches prioritized the latest CXR for simplicity~\cite{hayat2022medfuse, yao2024drfuse}, neglecting the temporal CXR context, while longitudinal fusion methods~\cite{zhang2023improving, lee2023learning} rely on rigid imputation or fixed time embeddings, lacking adaptive mechanisms to synchronize fine-grained cross-modal patterns. Thus, strategically integrating these progression dynamics to model cross-modal clinical trajectories remains underexplored. 

To address the aforementioned challenges, we propose \textit{\underline{Di}sease \underline{Pro}gression-Aware Clinical Prediction}, \dipro\footnote{The code is available at \url{https://github.com/Chenliu-svg/DiPro}.}, a novel framework that disentangles time-variant and time-invariant information from longitudinal CXRs and integrates these representations with EHR data across multiple temporal granularities. Specifically, our framework contains three modules:  \textbf{(1) Spatiotemporal Disentanglement}: We disentangle dynamic pathological changes from static anatomical structures in different spatial regions across consecutive CXRs. This separation helps reduce redundancy and prevents interference between features with different clinical semantics, allowing the model to focus on meaningful progression signals for more effective temporal representation learning.
\textbf{(2) Progression-Aware Enhancement}:  To improve the model’s sensitivity to progression direction, we reverse the order of CXR pairs and train the model to produce reversed dynamic features while keeping static features consistent. This strategy further separates the two types of features and emphasizes their distinct clinical semantics. \textbf{(3) Multimodal Fusion via Multiscale Alignment}: To fully synergize the multimodal features, we introduce a multi-grained fusion module that achieves alignment in multiple scales: the local fusion aligns consecutive CXR pairs with EHR data, while the global fusion aligns the semantics of the entire CXR and EHR sequences. This bridges fine-grained EHR-CXR interactions with global disease progression patterns.

Our contributions are summarized as follows:

\begin{itemize}[leftmargin=*, itemsep=3pt, parsep=0pt, topsep=1pt, partopsep=0pt]
\item We present a framework to disentangle dynamic and static information from longitudinal CXRs, with dedicated architectural constraints to capture representations in line with disease progressions.
\item We propose a multiscale multimodal fusion approach that facilitates multi-grained interactions between temporally misaligned CXR dynamics and EHR time-series data.
\item Extensive experiments demonstrate that \dipro achieves state-of-the-art performance on both disease progression modeling and ICU-related prediction tasks. Quantitative evaluation shows that the model aligns well with existing clinical knowledge.
\end{itemize}

\begin{figure}
    \centering
    \includegraphics[width=1.0\linewidth]{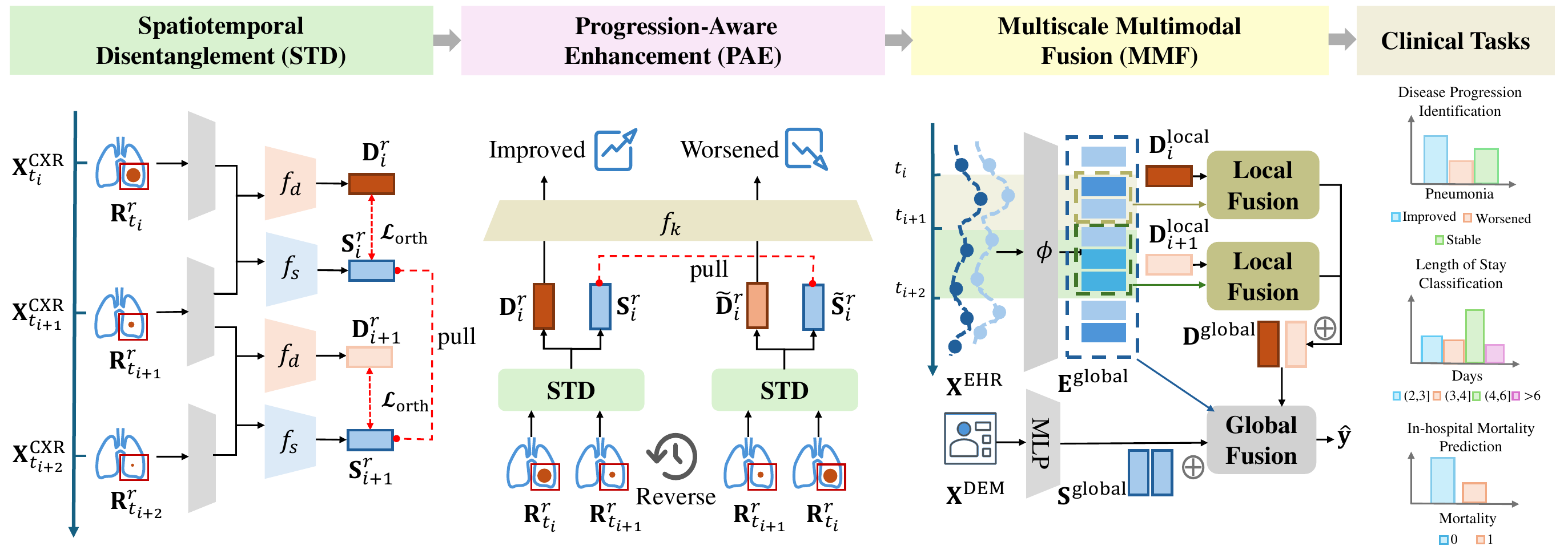}
    \caption{Overview of the \dipro~framework. The model comprises three main modules: (1) Spatiotemporal Disentanglement (STD) separates dynamic pathological features ($\mathbf{D}_{i}^{r}$) from static anatomical structures ($\mathbf{S}_{i}^{r}$) in region-level chest X-rays across time; (2) Progression-Aware Enhancement (PAE) strengthens the model’s understanding of progression direction by reversing CXR pair order and enforcing the reversed dynamic features $\widetilde{\mathbf{D}}_{i}^{r}$ to predict the reversed progression, while maintaining consistency in static features; (3) Multiscale Multimodal Fusion (MMF) integrates CXR features with temporally misaligned EHR data via local (interval-level) and global (sequence-level) fusion, enabling accurate predictions across multiple clinical tasks, including disease progression identification, length-of-stay classification, and in-hospital mortality prediction. }
    \label{fig:method}
\end{figure}

\section{The \dipro Approach}\label{sec:method}

To address the challenges of redundancy in longitudinal CXRs and temporal misalignment with EHR data, we propose \dipro, to systematically disentangle static and dynamic features from CXRs and align multimodal data across hierarchical timescales. The approach is motivated by two key observations: (1) disease progression in CXR sequences unfolds via region-localized pathological changes~\cite{wang2017chestx, pellegrini2023xplainer}, and (2) EHR and imaging data exhibit complementary dynamics at different temporal granularities. \dipro integrates three cohesive modules: (1) Spatiotemporal Disentanglement (STD) to isolate pathology-sensitive features from sequential CXRs, (2) Progression-Aware Enhancement (PAE) to enforce temporal consistency in learned dynamics, and (3) Multiscale Multimodal Fusion (MMF) to bridge local EHR-CXR interactions with global progression trends. \Cref{fig:method} illustrate the \dipro framework. We next detail each module.

\subsection{Notations and Preliminaries}
For each patient, let $\mathbf{X}^{\text{CXR}} = \{\mathbf{X}^{\text{CXR}}_{t_i}\}_{i=1}^T$ denote a set of $T$ CXR images during an ICU stay, where each $\mathbf{X}^{\text{CXR}}_{t_i} \in \mathbb{R}^{H \times W \times C}$ is the CXR image taken at time $t_i$. Each CXR image contains $R$ anatomical regions $\{\mathbf{R}_{t_i}^{r}\}_{r=1}^R$. Each consecutive image pair $(\mathbf{X}_{t_i}^{\text{CXR}}, \mathbf{X}_{t_{i+1}}^{\text{CXR}})$ is associated with a label set $\mathbf{Y}_i^{\text{CXR}} = \{y_{i}^{r,k}\}_{r,k=1}^{R,K}$ where $y_{i}^{r,k} \in \{-1, 0, 1\}$ indicates whether the progression status of disease $k$ in the $r$-th region has worsened, remained stable (no  change), or improved.
The whole EHR time serie recorded with $M$ timestamps is $ \mathbf{X}^{\text{EHR}} =  [\mathbf{x}_{t_0},\mathbf{x}_{t_{1}}, \dots, \mathbf{x}_{t_M}]$ with $\mathbf{x}_{t}\in \mathbb{R}^N$ being the $N$-dim variables recorded at time $t$. We use $\mathbf{X}^{\text{EHR}}$ to denote all available EHR time series.
Patient's demographic attributes are denoted as $\mathbf{X}^{\text{DEM}} \in \mathbb{R}^{P}$ where $P$ is the number of attributes. Our objective is to learn a mapping $f_\theta: (\mathbf{X}^{\text{CXR}}, \mathbf{X}^{\text{EHR}}, \mathbf{X}^{\text{DEM}}) \rightarrow \mathbf{y}$, that maps the integrated CXR, EHR time series, and demographic information to the final clinical outcome prediction $\mathbf{y}$.
\subsection{Spatiotemporal Disentanglement (STD)}
We propose a novel method to disentangle region-based time-variant (dynamic) and time-invariant (static) information from consecutive CXR image pairs. The disentanglement is motivated by their distinct clinical roles: dynamic features reflect disease progression, while static features capture patient-specific anatomical structures. Inspired by prior works demonstrating the effectiveness of feature disentanglement in representation learning~\cite{ liang2023quantifying, 
 zhu2021and}, our method targets more efficient and structured latent representations.
\paragraph{Feature extraction.}
For each anatomical region $\mathbf{R}_{t_i}^{r}$, we first use a shared pretrained ResNet-50~\cite{he2016deep} to decode it and obtain the region feature $\mathbf{F}_{t_i}^r \in \mathbb{R}^d$.
Then we concatenate the two consecutive feature vectors and then pass them through two separate projection heads, $f_s$ and $f_d$ for static and dynamic information extraction, respectively,: 
\begin{equation}\label{eq:stat-dyna}
\mathbf{S}_{i}^{r} = f_s([\mathbf{F}_{t_i}^{r} || \mathbf{F}_{t_{i+1}}^{r}]), \quad 
\mathbf{D}_{i}^{r} = f_d([\mathbf{F}_{t_i}^{r} || \mathbf{F}_{t_{i+1}}^{r}]),
\end{equation}
where $||$ denotes channel-wise concatenation. Here, $\mathbf{S}_{i}^{r}$ and $\mathbf{D}_{i}^{r}$ represent the static and dynamic features, respectively, for region $r$ at time step pair $(t_i, t_{i+1})$.

\paragraph{Orthogonal disentanglement loss.}
To encourage effective disentanglement between static and dynamic representations, we apply an orthogonal constraint~\cite{jiang2023understanding}. Specifically, we minimize the squared cosine similarity between static and dynamic features within each region and time pair:
\begin{equation}
\mathcal{L}_{\text{orth}} = \frac{1}{(T-1)R} \sum_{i=1}^{T-1} \sum_{r=1}^{R} \left( \text{sim} \left( \mathbf{S}_{i}^{r}, \mathbf{D}_{i}^{r} \right) \right)^2,
\end{equation}
where $\text{sim}(\cdot, \cdot)$ denotes the cosine similarity. This term ensures that the dynamic and static features are as unrelated as possible in the latent space, thereby promoting
better disentanglement between time-invariant and time-varying patterns.

\paragraph{Temporal consistency for static features.}
Anatomical structures are expected to remain stable over time in sequential CXR images~\cite{ren2022local, zhou2023learning}. To enforce temporal consistency of static features, we introduce a mean squared error (MSE) loss that encourages static features from consecutive time pairs to remain close:
\begin{equation}
\mathcal{L}_{\text{temp}} = \frac{1}{N} \sum_{r=1}^{R} \sum_{i=1}^{T-2} \left\| \mathbf{S}_{i}^{r} - \mathbf{S}_{i+1}^{r} \right\|_2^2,
\end{equation}
where $N = (T - 2) \times R$ is the number of consecutive static feature pairs considered. This constraint ensures that the learned static features remain stable and coherent over time.

\subsection{Progression-Aware Enhancement (PAE)}
Intuitively, reversing the order of a CXR pair should invert the progression direction while preserving static anatomical information. Hence, dynamic features should reflect this reversal, providing a more robust and interpretable representation of temporal change. To make dynamic features more sensitive to the direction of disease progression, we introduce a progression-aware enhancement. 

By reversing the input order of the region feature pair $(\mathbf{F}_{t_i}^{r},\mathbf{F}_{t_{i+1}}^{r})$, we obtain the dynamic and static features as:
\begin{equation}
    \widetilde{\mathbf{D}}_{i}^{r} = f_d([\mathbf{F}_{t_{i+1}}^{r} || \mathbf{F}_{t_i}^{r}]),\ \widetilde{\mathbf{S}}_{i}^{r} = f_s([\mathbf{F}_{t_{i+1}}^{r} || \mathbf{F}_{t_i}^{r}]) .
\end{equation}
We then feed both the original and reversed dynamic features into $K$ disease-specific progression classification heads $\{f_k\}_{k=1}^{K}$, where each head $f_k$ corresponds to a disease and predicts its progression status $y_{i}^{r,k}$ for the given region $r$.
Let $\widehat{y}^{r,k}_i$ be the predicted label from the original direction, and $\widetilde{y}_i^{r,k}$ be the prediction from the reversed input, i.e.,
$\widehat{y}_{i}^{r,k} = f_k(\mathbf{D}_{i}^{r}), \ \widetilde{y}_i^{r,k}= f_k(\widetilde{\mathbf{D}}_{i}^{r})$.
We expect that reversing the input order should generate a contrary prediction. Thus, we convert the label into $-y_{i}^{r,k}$ to indicate a reversed progression direction as the ground truth label.  

\paragraph{Training objective.}
For dynamic features, we supervise predictions using cross-entropy (CE) loss for both original and reversed progression prediction; For static features, we leverage MSE to encourage their consistency over the reversal.
\begin{equation}
    \mathcal{L}_{\text{PAE}} = \sum_{r=1}^{R} \sum_{k=1}^{K} \left[\text{CE}(\widehat{y}_{i}^{r,k}, y_{i}^{r,k}) + \text{CE}(\widetilde{y}_i^{r,k}, - y_{i}^{r,k}) \right] + \lambda_{\text{static}} \sum_{r=1}^{R} \left\| \mathbf{S}_{i}^{r} - \widetilde{\mathbf{S}}_{i}^{r} \right\|_2^2,
\end{equation}
where $\lambda_{\text{static}}$ is a hyperparameter. This objective encourages the model to encode progression-aware dynamic information and time-invariant anatomical information, improving the reliability of progression modeling across time.
\subsection{Multiscale Multimodal Fusion (MMF)}

To effectively integrate temporally disaligned patient data, we propose a multiscale fusion framework that combines visual and temporal features from longitudinal chest X-rays (CXRs) and electronic health records (EHRs). EHR signals are composed of dynamic time-series measurements and static demographics, while CXRs encode both time-varying imaging biomarkers and static anatomical traits. Our fusion proceeds in three stages: (1) local CXR-EHR fusion within each interval, (2) global hierarchical fusion, and (3) static feature integration and prediction.

\paragraph{Local CXR-EHR fusion within temporal intervals.}
We first encode the whole EHR time series $\mathbf{X}^{\text{EHR}}$ using a global Transformer encoder $\phi$~\cite{vaswani2017attention} to get the global representation of EHR: $\mathbf{E}^{\text{global}}  = 
\phi(\mathbf{X}^{\text{EHR}})$. Given a time interval $[t_i, t_{i+1}]$ from a consecutive image pair $(\mathbf{X}_{t_i}^{\text{CXR}}, \mathbf{X}_{t_{i+1}}^{\text{CXR}})$, to obtain EHR representations focused on the interval, we first define the time embedding for each EHR timestamp $t_j$ relative to the CXR time interval as:
$
\text{T}_{t_j} = f_{\text{TE}}\left([t_j - t_i,\ t_{i+1} - t_j,\ \sigma((t_j - t_i)(t_{i+1} - t_j))]\right),
$
where $\sigma(\cdot)$ is the sigmoid function to approximate the indicator of whether $t_j \in [t_i, t_{i+1}]$. These time embeddings are then stacked as $\mathbf{T}_{i} = [\text{TE}_{t_0}, \dots, \text{TE}_{t_M}]$. 
 we then apply cross-attention~\cite{vaswani2017attention} between the time embeddings and global EHR features with learnable parameters $\mathbf{W}_Q, \mathbf{W}_K$ and $\mathbf{W}_V$ using a center-focused attention mask:
\begin{equation}
\mathbf{E}_{i}^{\text{local}} =  \text{softmax}\left(\frac{\mathbf{Q} \mathbf{K}^\top}{\sqrt{d}} + \text{AttnMask}\right) \cdot \mathbf{V},\  \mathbf{Q} = \mathbf{W}_Q \mathbf{T}_{i},\ \mathbf{K} = \mathbf{W}_K \mathbf{E}^{\text{global}}, \ \mathbf{V} = \mathbf{W}_V \mathbf{E}^{\text{global}}.
\end{equation}
The attention mask is defined as:
\begin{equation}
\text{AttnMask}_{ij} = \begin{cases}
-\left|t_j - \frac{t_i + t_{i+1}}{2}\right|, & \text{if } t_j \in [t_i, t_{i+1}], \\
-\infty, & \text{otherwise}.
\end{cases}
\end{equation}
We then fuse dynamic CXR features $\mathbf{D}_{i}^{\text{local}} = \{ \mathbf{D}_{i}^{r} \}_{r =1}^R$ and local EHR representations $\mathbf{E}_{i}^{\text{local}}$ via a cross-attention layer. The keys and values are constructed by concatenating modality-specific projections of both inputs.
\begin{equation}
\mathbf{D}_{i}^{\text{fuse}} = \text{LayerNorm}(\text{CrossAttn}(\mathbf{D}_{i}^{\text{local}}, [\mathbf{E}_{i}^{\text{local}}|| \mathbf{D}_{i}^{\text{local}}]).
\end{equation}

\paragraph{Global hierarchical fusion}
We collect all locally fused features $\mathbf{D}^{\text{global}} = \{ \mathbf{D}_{i}^{\text{fuse}} \}_{i=1}^{T-1}$ across CXR intervals. We refine the global EHR representation by attending over $\mathbf{D}^{\text{global}}$:
\begin{equation}
    \mathbf{H}^{\text{global}} = \text{LayerNorm}(\text{CrossAttn}(\mathbf{E}^{\text{global}}, \mathbf{D}^{\text{global}})).
\end{equation}
We include an additional self-attention layer~\cite{vaswani2017attention} to further enhance global interactions between the two modalities. The resulting enriched sequence then serves as the query in the final cross-attention mechanism with the static features for prediction.

\paragraph{Final static fusion and prediction.}
We first embed demographic information $\mathbf{X}^{\text{DEM}}$ via an MLP, then concatenate it with static CXR features $\mathbf{S}^{\text{global}} = \{ \mathbf{S}_{i} \}_{i=1}^T$:
\begin{equation}
\mathbf{H}^{\text{static}} = [\mathbf{S}^{\text{global}} \,||\, \text{MLP}(\mathbf{X}^{\text{DEM}})].
\end{equation}

A fusion module $\Phi$, composed of a cross-attention layer followed by prediction heads, integrates dynamic and static features to generate predictions:
\begin{equation}
\widehat{\mathbf{y}} = \Phi([\mathbf{D}^{\text{global}} \,||\, \mathbf{H}^{\text{global}}], \mathbf{H}^{\text{static}}).
\end{equation}

The final training loss combines cross-entropy with all auxiliary objectives:
\begin{equation}
\mathcal{L}= \lambda_{\text{pred}} \cdot \text{CE}(\widehat{\mathbf{y}}, \mathbf{y}) + \lambda_{\text{orth}} \mathcal{L}_{\text{orth}} + \lambda_{\text{temp}} \mathcal{L}_{\text{temp}} + \lambda_{\text{PAE}} \mathcal{L}_{\text{PAE}},
\end{equation}
where $\lambda_{\text{pred}}$, $\lambda_{\text{orth}}$, $\lambda_{\text{temp}}$, and $\lambda_{\text{PAE}}$ are hyperparameters. Details of the model architecture, hyperparameters, and training procedure are provided in~\Cref{app:arch_train}.

\section{Experiments}\label{sec:experiment}
\subsection{Experiment Setting}
\paragraph{Datasets.} We evaluated \dipro on the large-scale, public dataset, MIMIC~\cite{johnson2023mimic}, which contains de-identified health data of intensive care unit (ICU) admissions. Our study leveraged three derived datasets from the MIMIC ecosystem: (1) \textbf{MIMIC-IV}~\cite{johnson2023mimic} provides electronic health records (EHR) including demographic information and time-series physiological measurements per ICU stay; (2) \textbf{MIMIC-CXR}~\cite{johnson2019mimic} contains sequential chest radiographs during ICU hospitalizations; and (3) \textbf{Chest ImaGenome}~\cite{wu2021chest} augments imaging with fine-grained annotations: bounding boxes for anatomical regions and localized change labels (improved, worsened, or no change) between consecutive CXRs.
To facilitate longitudinal analysis, we selected ICU stays with $\geq$ 2 CXRs to track disease progression. From MIMIC-IV, we extracted EHR data consisting of 7 demographic variables and 38 physiological time-series variables, including vital signs and laboratory results. 

\paragraph{Clinical tasks and evaluation metrics.}
We evaluate the performance of \dipro on two types of clinical tasks to demonstrate the advantages of our multimodal framework:

(1)\textit{Disease progression identification.} Given two consecutive CXRs, the task is to predict the disease progression status (improved, worsened, or no change) for seven common thoracic conditions: 
atelectasis, enlarged cardiac silhouette, consolidation, pulmonary edema, lung opacity, pleural effusion, and pneumonia. 
Following~\citet{karwande2022chexrelnet}, we derive the progression label of each disease for a CXR pair from the progression labels of annotated regions in the Chest ImaGenome dataset. 
Notably, our work is the first to integrate EHR data for this task. 
Specifically, we extract the EHR data recorded within the time interval of two CXRs and integrate them for a richer context for progression. We report macro-precision, macro-recall, macro-F1 score, AUPRC and AUROC following prior works~\cite{zhu2024symptom, karwande2022chexrelnet, mbakwe2023hierarchical}.

(2)\textit{General ICU prediction.} We consider two clinically vital tasks: In-hospital mortality prediction and ICU length of stay prediction. 
Both tasks focus on forecasting patient outcomes leveraging multimodal data: EHR time series and sequential CXRs collected during the first 48 hours after ICU admission. 
The \textit{In-hospital mortality prediction} task is a binary classification problem: it predicts patient mortality prior to hospital discharge. We evaluate performance using AUROC and AUPRC, following~\cite{zhang2023improving, hayat2022medfuse}.  
\textit{Length of stay prediction} task aims to estimate patient ICU stay duration. We frame this as a multi-class classification problem by discretizing stay duration into four intervals: [2, 3), [3, 4), [4, 6), and $\geq 6$ days. The counting starts from 2 days as we are using ICU stays longer than 48 hours. Following~\cite{harutyunyan2019multitask, xu2018raim, song2018attend}, model performance is evaluated using Cohen's kappa and accuracy.

All experiments are conducted with three random seeds, with results reported as the mean $\pm$ standard deviation across independent runs. Details on cohort selection, label prevalence, data statistics and data processing procedures for each task are provided in~\Cref{app:Data-Preprocessing}.
\paragraph{Baselines.} We compare \dipro with the following three types of baselines:

(1) \textbf{Sequential CXR disease progression specialists (unimodal)}: 
\textit{CheXRelNet}~\cite{karwande2022chexrelnet} combines local and global visual features with anatomical dependencies to model longitudinal disease changes. 
\textit{CheXRelFormer}~\cite{mbakwe2023hierarchical} adopts hierarchical Siamese Transformer to capture multi-level feature discrepancies across CXR images. 
\textit{SDPL}~\cite{zhu2024symptom} learns symptom-aware embeddings to extract and compare condition-specific features from two radiographs. 

(2) \textbf{Longitudinal multimodal specialists}: 
\textit{UTDE}~\cite{zhang2023improving} models asynchronous longitudinal data via a gated attention-based imputation framework. 
\textit{UMSE}~\cite{lee2023learning} uses triplet-structured set embeddings and a modality-aware attention mechanism to handle missing data and fuse multiple modalities.

(3) \textbf{Clinical multimodal fusion specialists}: 
\textit{MedFuse}~\cite{hayat2022medfuse} introduces an LSTM-based module for both uni-modal and multimodal input. 
\textit{DrFuse}~\cite{yao2024drfuse} disentangles modality-shared and modality-specific features, and utilizes disease-wise attention for effective fusion.
Both models are designed to take the last available CXR for modality input, we extend them to the setting of multiple CXRs with minimal architectural modification. Details of all the baseline models are provided in Appendix~\ref{app:baselines}.

\subsection{Prediction Performance}

\paragraph{\dipro excels in modeling disease progression in sequential CXRs.}  
\Cref{tab:dp_results} reports the mean performance across seven disease progression identification tasks (per-disease results in~\Cref{tab:dp_decompose}). Compared to CXR-based progression models using only unimodal sequential CXRs, \dipro achieves relative improvements of 15.3\% in F1 and 12.2\% in AUPRC over the state-of-the-art SDPL~\cite{zhu2024symptom}. This suggests that explicitly disentangling disease dynamics from static anatomical structures across CXR pairs reduces redundancy and enables more effective progression modeling. Compared to CheXRelNet~\cite{karwande2022chexrelnet}, which models region-disease co-occurrence via graphs, \dipro enhances the disentangled regional progression dynamics using a tailored PAE module, offering a more targeted mechanism to capture disease-region progression. A broader baseline comparison with large vision–language models is presented in~\Cref{tab:vlm_comparison}.

\begin{table}[ht!]\small
    \centering
    \caption{\textbf{Performance Comparison on Disease Progression Identification Tasks.}
This table reports the macro-average performance ($\pm$ standard deviation) of various unimodal and multimodal methods across seven disease progression identification tasks. \dipro achieves the best results in both unimodal and multimodal settings, indicating its effectiveness in modeling disease progression from sequential CXRs and its strength in longitudinal multimodal fusion. Detailed results are provided in~\Cref{tab:dp_results-appendix}. Per-disease results are provided in~\Cref{tab:dp_decompose}.  (Numbers in bold indicate the best performance in each column, and those underlined represent the best-performing baseline.)}
    \label{tab:dp_results}
\begin{tabular}{lccccccc}
\toprule
Method & Precision & Recall & F1 & AUPRC & AUROC  \\
\midrule

\multicolumn{6}{c}{\textbf{Unimodal Methods (CXR)}}\\
\hdashline
   
CheXRelNet~\cite{karwande2022chexrelnet}
    & 0.395$\pm$0.015   & 0.392$\pm$0.010     & 0.389$\pm$0.010    & 0.394$\pm$0.010    & 0.574$\pm$0.011    
 \\

CheXRelFormer~\cite{mbakwe2023hierarchical}
   & 0.389$\pm$0.044     & 0.379$\pm$0.033     & 0.354$\pm$0.032     & 0.372$\pm$0.023     & 0.551$\pm$0.041      \\

SDPL~\cite{zhu2024symptom}
     & \underline{0.408$\pm$0.006}     & \underline{0.406$\pm$0.020}     & \underline{0.393$\pm$0.010}     & \underline{0.417$\pm$0.032}     & \underline{0.609$\pm$0.031}    
\\

\dipro(ours)
 & \textbf{0.475$\pm$0.004}     & \textbf{0.452$\pm$0.011}     & \textbf{0.453$\pm$0.009}     & \textbf{0.468$\pm$0.013 }    &\textbf{ 0.651$\pm$0.016  }    \\

\midrule
\midrule

\multicolumn{6}{c}{\textbf{Multimodal Methods}}\\
\hdashline
UTDE~\cite{zhang2023improving}
&  \underline{0.481$\pm$0.017}    & \underline{0.462$\pm$0.002}    & \underline{0.449$\pm$0.005}    & \underline{0.472$\pm$0.014 }   & \underline{0.659$\pm$0.011}\\

UMSE~\cite{lee2023learning} &  0.353$\pm$0.011     & 0.361$\pm$0.009     & 0.352$\pm$0.013     & 0.364$\pm$0.006     & 0.544$\pm$0.004   \\

MedFuse~\cite{hayat2022medfuse} & 0.423$\pm$0.049     & 0.413$\pm$0.045     & 0.409$\pm$0.042     & 0.422$\pm$0.040       & 0.530$\pm$0.030 \\
DrFuse~\cite{yao2024drfuse}
& 0.442$\pm$0.009     & 0.461$\pm$0.007   & 0.429$\pm$0.010     & 0.438$\pm$0.003     & 0.628$\pm$0.002     \\
  
 \dipro(ours)& \textbf{0.484$\pm$0.008}     & \textbf{0.471$\pm$0.024}     & \textbf{0.466$\pm$0.018}     & \textbf{0.478$\pm$0.018 }    & \textbf{0.664$\pm$0.013}    \\
\bottomrule
\end{tabular}
\end{table}

\paragraph{\dipro excels in longitudinal multimodal fusion.}  
As shown in the \textit{Multimodal Methods} block of~\Cref{tab:dp_results}, adding EHR to unimodal \dipro improves performance (relative increase of 2.9\% in F1 and 2.1\% in AUPRC). This confirms \dipro's ability to effectively leverage complementary EHR features for disease progression prediction. Furthermore, \dipro outperforms all baselines, whether using the last available CXR or longitudinal CXR in multimodal settings, or using unimodal EHR, across all tasks, including disease progression identification (\Cref{tab:dp_results}) and general ICU prediction (\Cref{tab:general-icu-results}). Specifically, it achieves relative gains of 3.8\% in F1 score (disease progression), 2.8\% in AUPRC (mortality prediction), and 3.0\% in accuracy (length of stay) compared to the respective best-performing baselines. Notably, \dipro consistently outperforms longitudinal multimodal specialists like UTDE~\cite{zhang2023improving} and UMSE~\cite{lee2023learning}. These results highlight the superiority of \dipro's multiscale fusion strategy in addressing temporal misalignment compared to unimodal imputation or unified time-embedding methods.

\begin{table}\small
    \centering

\caption{\textbf{Performance Comparison on General ICU Prediction Tasks.}
This table presents the average performance ($\pm$ standard deviation) on two general ICU prediction tasks: mortality (AUPRC, AUROC) and length of stay (Kappa, ACC). Results are reported for both ``Last'' and ``Long.'' CXR settings. The ``Long.'' setting incorporates both longitudinal CXR and EHR data, whereas the ``Last'' setting uses only the most recent CXR together with EHR data. \dipro demonstrates superior performance across both settings, highlighting its effectiveness in longitudinal multimodal fusion. Note that in the ``no-CXR'' setting (second-to-last row), \dipro effectively reduces to a Transformer-based~\cite{vaswani2017attention} EHR encoder. More results are provided in~\Cref{tab:gen-icu-performance-appendix}.}
    \label{tab:general-icu-results}
    \begin{tabular}{lcccccc}
    \toprule
     & \multicolumn{2}{c}{CXR Used} & \multicolumn{2}{c}{Mortality} & \multicolumn{2}{c}{Length of Stay} \\\cmidrule(lr){2-3}\cmidrule(lr){4-5}\cmidrule(lr){6-7}
     Method & Last & Long.   & AUPRC       & AUROC  & Kappa & ACC      \\\midrule
     
     UTDE~\cite{zhang2023improving} &\coloredcheckmark &  &  0.717$\pm$0.019     & 0.887$\pm$0.004 &  0.160$\pm$0.016  & 0.381$\pm$0.013   \\
      & &\coloredcheckmark  & 0.710$\pm$0.019     & 0.887$\pm$0.012 & 0.195$\pm$0.031  & 0.400$\pm$0.021
 \\
     UMSE~\cite{lee2023learning}    &\coloredcheckmark &  & \underline{0.722$\pm$0.039}    & \underline{0.896$\pm$0.012}  & 0.217$\pm$0.013   & 0.419$\pm$0.010  \\
     & & \coloredcheckmark& 0.712$\pm$0.028     & 
     0.891$\pm$0.011 & 0.204$\pm$0.019 & 0.410$\pm$0.013 \\
    \midrule
     \multirow{2}{*}{MedFuse~\cite{hayat2022medfuse}}
      &\coloredcheckmark &  & 0.686$\pm$0.018     & 0.869$\pm$0.011 & 0.213$\pm$0.012  & 0.413$\pm$0.004 
 \\
     & & \coloredcheckmark & 0.716$\pm$0.018     & 0.881$\pm$0.005 &0.210$\pm$0.039   & 0.412$\pm$0.027
 \\
     \multirow{2}{*}{DrFuse~\cite{yao2024drfuse}}
       &\coloredcheckmark &   & 0.709$\pm$0.012     & 0.865$\pm$0.014 & 0.114$\pm$0.048  & 0.338$\pm$0.041
 \\
     & & \coloredcheckmark& 0.684$\pm$0.008     & 0.854$\pm$0.017 & 0.142$\pm$0.014 & 0.360$\pm$0.011 
  \\
    \midrule
     \multirow{2}{*}{\dipro (Ours)}
      & & & 0.712$\pm$0.009     & 0.885$\pm$0.003   & \underline{0.226$\pm$0.019}  & \underline{0.427$\pm$0.014}  \\
    & &\coloredcheckmark & \textbf{0.742$\pm$0.003}     & \textbf{0.897$\pm$0.002}   & \textbf{0.248$\pm$0.008} & \textbf{0.440$\pm$0.007}\\
    \bottomrule
    \end{tabular}
\end{table}

\paragraph{\dipro alleviates redundancy and misalignment in sequential CXRs.}  
\Cref{tab:general-icu-results} presents a comparison of multimodal models that integrate EHR data with either the most recent CXR (\textit{Last}) or longitudinal CXRs (\textit{Long.}). For UMSE~\cite{lee2023learning}, using longitudinal CXRs do not consistently improve general ICU prediction performance compared to using only the latest image, suggesting that unified time embeddings and bottleneck attention inadequately address temporal redundancy in sequential imaging data.
Similarly, DrFuse~\cite{yao2024drfuse} experiences a slight performance drop in mortality prediction when naively concatenating sequential CXR features, highlighting the limitations of direct aggregation without temporal alignment.
Alternatively, \dipro explicitly disentangles disease progression dynamics from time-invariant CXR features, reducing redundancy. It aligns multimodal trajectories by fusing progression-aware features with EHR data at both interval and sequence levels, enabling more efficient use of longitudinal data. This approach yields superior performance across tasks. 

\begin{figure}
    \centering
    \begin{subfigure}[t]{0.4\linewidth}
    \centering
    \includegraphics[width=\linewidth]{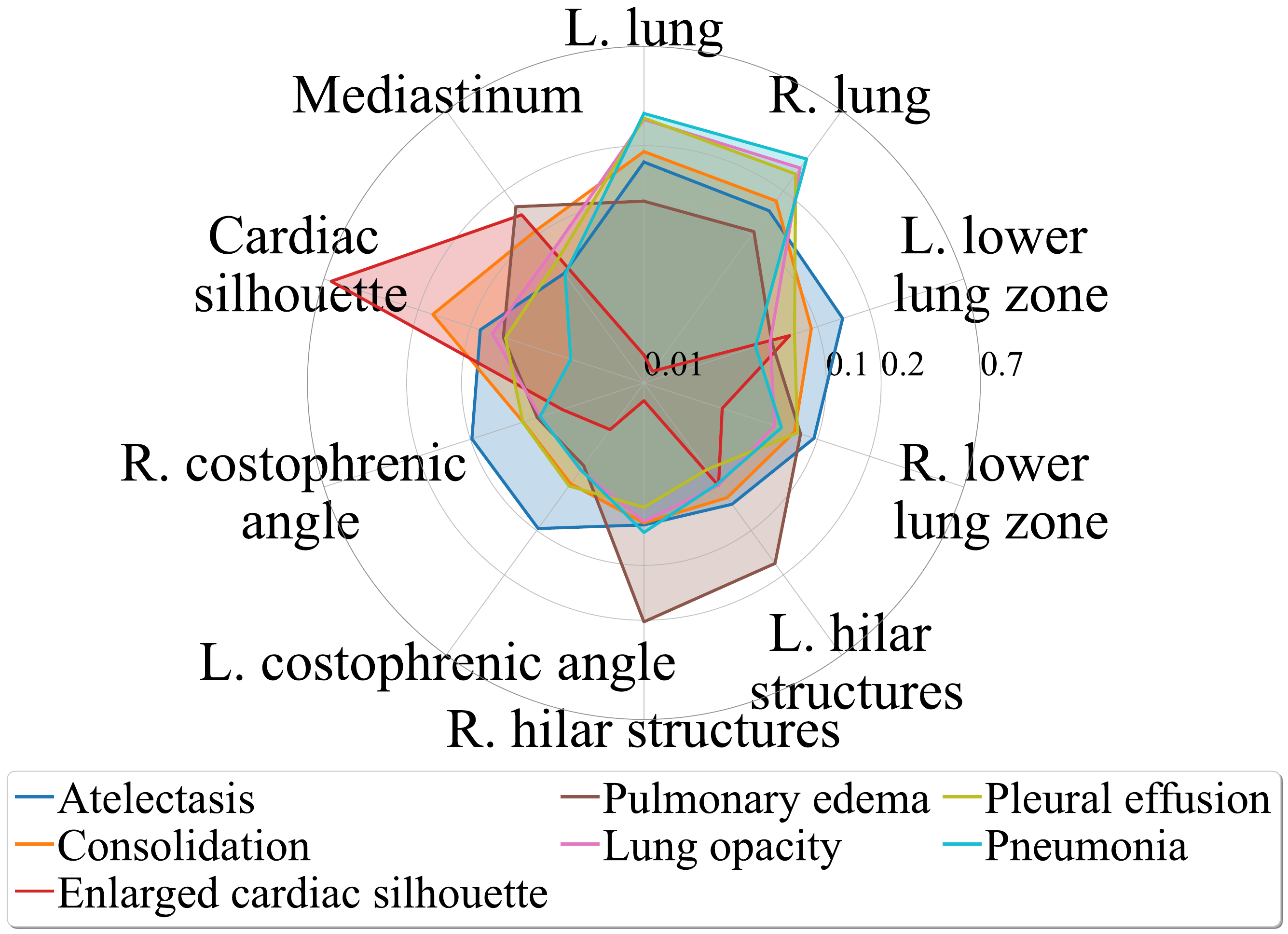}
    \caption{Disease Progression Prediction}
    \label{fig:dp_attn}
    \end{subfigure}
    \begin{subfigure}[t]{0.59\linewidth}
    \centering
    \includegraphics[width=\linewidth]{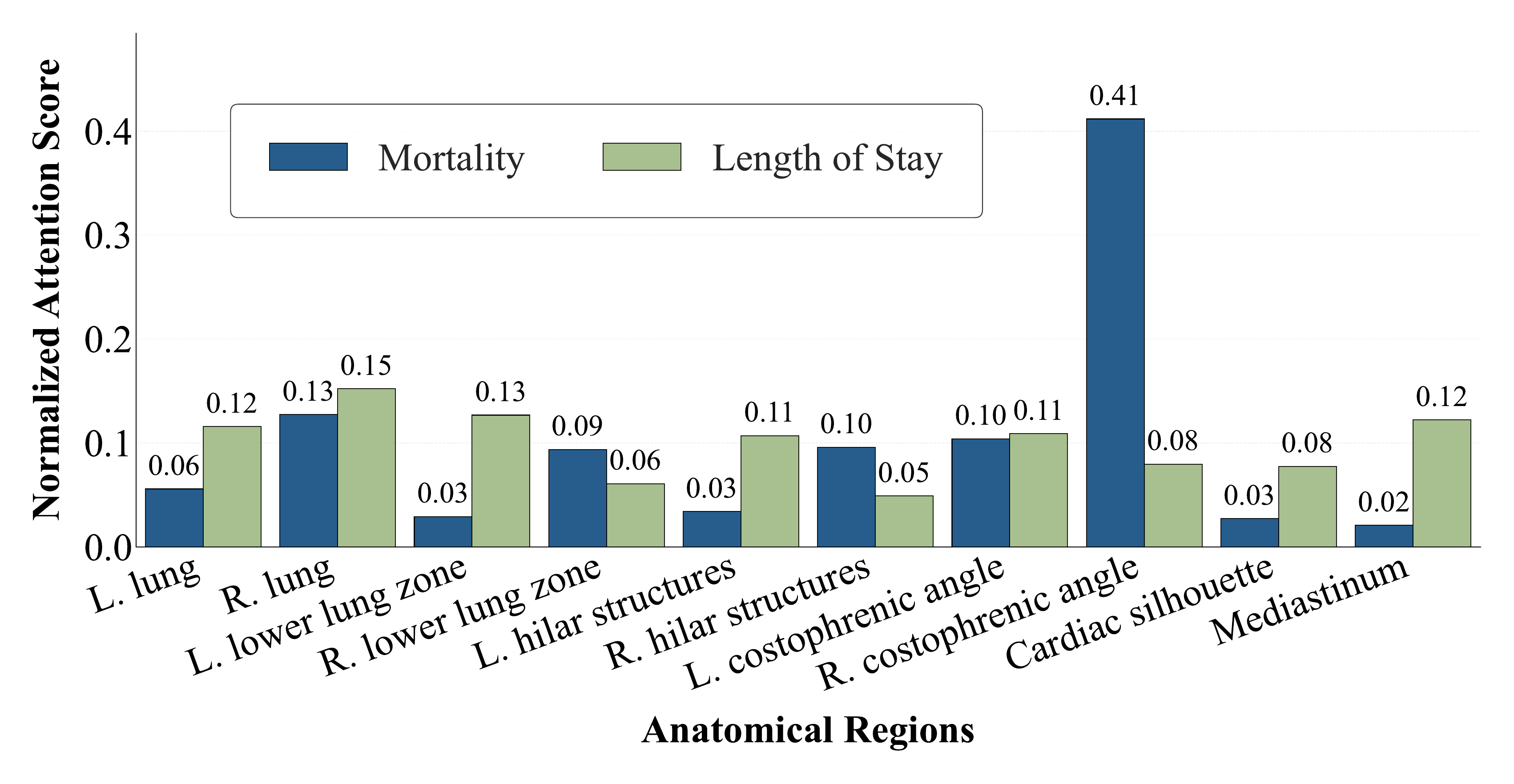}
    \caption{Mortality and Length-of-stay Prediction}
    \label{fig:icu_attn}
\end{subfigure}
    \caption{Averaged attention weights of CXR regions in different downstream tasks. The radial axis in (a) is log-scaled to enhance distribution visibility. Mean attention weights across CXR regions reveal \dipro's clinical alignment: (a) overlapping distributions for pneumonia, lung opacity, and pleural effusion reflect shared pathologies, while (b) ICU tasks show divergent patterns: higher weights for right-sided regions in mortality (linked to higher risk) versus diffuse attention in length-of-stay (reflecting multifactorial ICU conditions).}
\end{figure}

\paragraph{\dipro echoes with clinical knowledge.}
To analyze the anatomical basis of \dipro's decision-making, we visualize the normalized attention weight of CXR regions of each disease in the disease progression task. As shown in~\Cref{fig:dp_attn}, \dipro demonstrates high attention scores on the cardiac silhouette when identifying cardiomegaly, which aligns with the radiographic diagnostic criteria where cardiothoracic ratio (CTR) $>0.5$ on posteroanterior chest radiographs indicates cardiac enlargement~\cite{agrawal2023segmentation, yanar2023clinical}. Similarly, the model highlights hilar structures for pulmonary edema detection, corroborating the pathophysiological mechanism that pulmonary venous congestion in left ventricular failure manifests as perihilar vascular redistribution and interstitial edema~\cite{wilkins2006respiratory}. Notably, the radar plot reveals overlapping attention weight distributions for pneumonia, lung opacity, and pleural effusion, suggesting shared radiographic features due to common pathways~\cite{cilloniz2018epidemiology, elmukhtar2024diagnostic, marcdante2014nelson}. This suggests that \dipro captures clinically meaningful correlations in radiographic patterns. 

In~\Cref{fig:icu_attn}, we further analyze \dipro's normalized regional attention weights for two ICU prediction tasks. For in-hospital mortality prediction, \dipro assigns notably high attention to the right costophrenic angle, a region clinically associated with pleural effusions and lower lobe pathologies, both of which are common in critically ill patients and have been linked to increased mortality risk~\cite{guideline2002non, vincent2002epidemiology}. Furthermore, the model consistently prioritizes right-sided anatomical structures over left ones (e.g., right lung: 0.13 vs. left: 0.06; right hilar region: 0.10 vs. left: 0.03). This pattern aligns with the predominance of right-sided pulmonary complications (e.g., aspiration pneumonia and pleural effusion), linked to increased risk of mortality~\cite{lohan2019imaging, bediwy2023pleural, maslove2013diagnosis}. For length-of-stay prediction, however, the model exhibits a more distributed attention pattern across multiple thoracic regions, including bilateral lungs, hilar structures, and mediastinum. This scattered pattern suggests that predicting the length-of-stay requires a broader view of radiographic features, which is consistent with the understanding that hospital stay duration aggregates diverse and multifactorial conditions, such as pulmonary congestion, atelectasis, and cardiomegaly~\cite{wright2003factors, gruenberg2006factors}.

\subsection{Ablation Study}
To better each component's contribution in \dipro, we ablate key modules (results in~\Cref{tab:ablation_study}). The variant ``A1'' replaces MMF with a simple fusion strategy that concatenates CXR and EHR features, followed by a multi-head self-attention layer. The variant ``A2'' removes the PAE module. The variant ``A3'' removes both MMF and PAE, leaving only a basic aggregation using self-attention. The ``A4'' variant removes the STD module and disables MMF and PAE, reducing \dipro to a plain encode-concatenate-attention baseline. 

As shown in~\Cref{tab:ablation_study}, removing any component of \dipro results in performance drop, underscoring the necessity of each module. Notably, contributions vary across tasks.
Different components contribute variably across tasks. For disease progression identification, the STD module is critical, yielding relatively F1 improvements  (``A3'' vs. ``A4'')  by 21.3\% through disentangling progression dynamics from static anatomical features. The PAE module further enhances performance (7.6\% relative F1 gain, \dipro vs. ``A2''), highlighting the benefit of progression-aware feature enhancement. However, the MMF module contributes less to the same task, possibly due to the constraint of using only EHRs linked to CXR pairs, restricting the multiscale fusion capability. Conversely, for general ICU prediction, MMF notably improves performance, highlighting the value of a well-designed fusion strategy for handling longitudinal multimodal data.Interestingly, incorporating the STD module into a simple encode-concatenate-attention baseline results in performance degradation (``A3'' vs. ``A4''), suggesting that disentanglement alone, without dedicated dynamic/static modeling and multiscale fusion, is insufficient for complex prediction tasks. More metrics can be found in~\Cref{tab:ablation_study_disease_pro_appendix,tab:ablation_study_los_appendix}.

\begin{table}
\centering
\caption{\textbf{Results of the ablation study.} 
This table presents the results of ablating major modules to assess their contribution to overall performance. 
Variants ``A1''–``A4'' correspond to variants of \dipro with progressively removed modules, while ``\dipro-'' denotes the variant using automated bounding boxes generated by the RGRG model~\cite{tanida2023interactive} instead of Chest ImaGenome annotations.
}
\label{tab:ablation_study}
\begin{tabular}{lcccccc}
\toprule
&\multicolumn{3}{c}{Components} & Disease Progression & Mortality&Length of stay\\
 \cmidrule(lr){2-4} \cmidrule(lr){5-5} \cmidrule(lr){6-6} \cmidrule(lr){7-7}
ID &  STD & PAE & MMF  & F1 &AUPRC  & ACC  \\
 \midrule
\dipro&\coloredcheckmark & \coloredcheckmark &  \coloredcheckmark & \textbf{0.466$\pm$0.018} & \textbf{0.742$\pm$0.003} & \textbf{0.440$\pm$0.007} \\
\hdashline
 \dipro-&\coloredcheckmark & \coloredcheckmark &  \coloredcheckmark& 0.457$\pm$0.010 & 0.736$\pm$0.021 & 0.430$\pm$0.006 \\
A1&\coloredcheckmark & \coloredcheckmark & \xmark  & 0.460$\pm$0.014 & 0.724$\pm$0.015     & 0.416$\pm$0.027 \\
 A2&\coloredcheckmark & \xmark & \coloredcheckmark  & 0.433$\pm$0.017 & 0.730$\pm$0.029     & 0.432$\pm$0.018 \\

A3&\coloredcheckmark & \xmark & \xmark  & 0.439$\pm$0.007  & 0.694$\pm$0.016   & 0.404$\pm$0.014 \\
A4& \xmark & \xmark & \xmark & 0.362$\pm$0.016  & 0.721$\pm$0.036      & 0.425$\pm$0.031\\ 
\bottomrule
\end{tabular}
\end{table}

\paragraph{Robustness with Automated Bounding Boxes}

To assess \dipro's robustness to automated region annotations, we conducted an ablation study replacing Chest ImaGenome bounding boxes with those generated by the automated region detection model from RGRG~\cite{tanida2023interactive},denoted as ``\dipro-'' in~\Cref{tab:ablation_study}. While using automated bounding boxes results in a slight performance drop compared to curated annotations, \dipro consistently outperforms all baselines across disease progression (0.449$\pm$0.005 in F1), mortality (0.722$\pm$0.039 in AUPRC), and length-of-stay (0.427$\pm$0.014 in ACC) tasks. This demonstrates that \dipro remains effective even in the absence of manually curated labels, supporting its applicability to datasets lacking fine-grained anatomical annotations.  While more accurate anatomical annotations can improve prediction performance, \dipro still achieves strong and generalizable results even with fully automated region proposals.

The ablation study on loss penalties is presented in~\Cref{tab:loss_ablation}, the final selected penalty weights for each prediction task are summarized in~\Cref{tab:penalty_weights}, and the robustness analysis under missing EHR data is reported in~\Cref{tab:missing_los,tab:missing_mortality}.

\section{Related Work}

\paragraph{Modeling disease progression in sequential CXRs.}
Recent years have seen growing interest in leveraging longitudinal CXRs for clinical prediction, as they are routinely used to monitor disease progression and naturally provide sequential imaging data \cite{zhu2024symptom, karwande2022chexrelnet, mbakwe2023hierarchical,wang2024hergen, bannur2023learning, serra2023controllable, zhuang2025advancing, eshraghi2024representation}. Most methods focus on capturing temporal differences using deep learning architectures. For instance, \citet{karwande2022chexrelnet} used a graph attention network to model region-level temporal changes, while \citet{eshraghi2024representation} adopted a Transformer-based detection model for localized progression signals. \citet{wang2024hergen} introduced time-aware causal attention, and \citet{mbakwe2023hierarchical} proposed a hierarchical Transformer for multi-scale comparison. Other approaches enhance clinical relevance via auxiliary tasks, such as symptom prediction \cite{zhu2024symptom} or spatiotemporal contrastive learning with radiology reports \cite{liu2025enhanced}. Despite these efforts, effectively addressing redundancy in sequential CXR modeling remains a fundamental challenge. 
\paragraph{Leveraging multimodal data for clinical prediction.}
Multimodal data offers rich temporal and semantic information for clinical prediction tasks \cite{zhuang2025advancing, kmetzsch2022disease, susman2024longitudinal, li2023longitudinal, warner2024multimodal, feng2025asynchronous, lian2025efficient}. Several methods combine the latest CXR with EHR data to improve performance using sophisticated fusion strategies \cite{hayat2022medfuse, yao2024drfuse, feng2024unified, yao2024addressing}. More recent efforts target the challenges of heterogeneous, misaligned longitudinal data. For example, \citet{li2023longitudinal} applied ICA to extract latent EHR signals and aligned them with CT scans using time-aware Transformers. \citet{susman2024longitudinal} proposed ensemble models to integrate multimodal sequences and highlight salient features for dementia prediction. Others address temporal irregularity directly: \citet{lee2023learning} introduced unified time embeddings and modality-aware attention, while \citet{zhang2023improving} imputed sparse clinical notes with temporal attention and fused them with multivariate time series. Yet, the core challenge of capturing disease progression across misaligned modalities remains underexplored \cite{zhuang2025advancing, yao2024addressing, zhao2021longitudinal}, limiting our ability to fully leverage cross-modal synergy in clinical contexts.
\section{Impacts and Limitations}
\label{sec:impact_limitation}
\dipro~advances multimodal disease progression modeling through its efficient integration of regional progression-aware feature disentanglement and multi-timescale alignment. The approach demonstrates significant potential for generalization to asynchronous clinical workflows, such as Alzheimer's disease monitoring using longitudinal MRI/PET scans with cognitive test records, or heart failure progression tracking using periodic echocardiograms with continuous vital signs. However, the current implementation relies on anatomical annotations (bounding boxes) to localize progression-specific features. While effective, this requirement may limit scalability in practice. Future work could reduce dependency on manual labels by adopting emerging segmentation tools, such as medical SAMs or weakly supervised localization methods~\cite{S-SAM}, which can better align the framework with real-world clinical workflows. Meanwhile, Our study excludes visits with only a single CXR. This selection may introduce sampling bias and reduce the overall cohort size. However, it is necessary for modeling disease progression between consecutive CXRs, as the longitudinal task inherently requires at least two images per patient. Developing methods that can handle single-CXR visits remains an important direction for future work.

\section{Conclusion}
In this paper, we propose \dipro, a novel framework that tackles critical challenges in fusing longitudinal multimodal data for clinical tasks: redundancy in sequential CXRs and temporal misalignment across modalities. By explicitly disentangling disease progression dynamics from static anatomical features via dedicated constraints, \dipro extracts clinical meaningful and discriminative dynamic/static patterns. To further enhance temporal alignment, we propose a  multiscale multimodal fusion strategy that bridges CXR-derived progression features with EHR time-series data through interval-wise and full-sequence-level interactions. Extensive experiments demonstrate that \dipro achieves state-of-the-art performance on both disease progression identification and general ICU prediction tasks, while providing interpretability consistent with clinical understanding.

\section*{Acknowledgments and Disclosure of Funding}
This work is partially supported by an Innovation and Technology Fund of Hong Kong Innovation and Technology Commission (project no. ITS/202/23), a Collaborative Research Fund of Hong Kong Research Grants Council (project no. C5055-24G), the National Natural Science Foundation of China~(62302413), the Health and Medical Research Fund~(23220312), the General Research Fund RGC/HKBU12202621 from the Research Grant Council, and the Research Matching Grant Scheme RMGS2021\_8\_06 from the Hong Kong Government.





\newpage
\appendix
\section{Experiment Details}
\label{app:exp}

\subsection{Details of Data Preprocessing}
\label{app:Data-Preprocessing}
\paragraph{EHR Data Preprocessing}
We adapted the EHR data processing pipeline from~\cite{hayat2022medfuse}, modifying the sampling frequency from 2-hour to 1-hour intervals. We incorporate vital signs, laboratory measurements, and clinical scores, resulting in a total of 38 clinical time-series variables, including alanine aminotransferase, albumin, alkaline phosphate, anion gap, asparate aminotransferase, bicarbonate, bilirubin, blood urea nitrogen, chloride, creatinine, diastolic blood pressure, fraction inspired oxygen, Glasgow coma scale (eye opening, motor response, and verbal response), glucose, heart rate, height, hematocrit, hemoglobin, magnesium, mean blood pressure, oxygen saturation, partial pressure of carbon dioxide, partial thromboplastin time, platelets, positive end-expiratory pressure, potassium, prothrombin time, respiratory rate, sodium, systolic blood pressure, temperature, troponin-T, urine output, daily weight, white blood cell count, and pH. Following~\cite{hayat2022medfuse}, we applied identical discretization and standardization procedures. Demographic data including age, height, admission weight, gender, race, language, and marital status were also incorporated.

\paragraph{CXR Preprocessing} CXR studies were temporally aligned with corresponding ICU stays to construct longitudinal imaging sequences. We retained only AP view images, matched them to ICU stays based on study timestamps, and excluded outlier admissions using length-of-stay filtering. Each matched CXR was restricted to those with available bounding-box annotations in Chest ImaGenome.  Meanwhile, in this study, we do not perform explicit longitudinal registration of CXR images. To mitigate potential positional misalignment over time, we follow the preprocessing strategy of CheXRelNet~\cite{karwande2022chexrelnet}: we apply cropping based on anatomical bounding boxes from Chest ImaGenome dataset. This preprocessing allows us to focus on capturing the semantic-level disease progression within consistent anatomical structure, rather than relying on pixel-level alignment. As a result, our model is designed to be more robust to positional variability across different time points.

\paragraph{Data Splitting and Statistics}
Our analysis focused on ICU stays containing at least two CXRs. The dataset was partitioned into training (70\%), validation (10\%), and test sets (20\%) at the subject level to prevent data leakage.~\Cref{tab:stats} summarizes the sample counts for each task, while~\Cref{tab:disease_progression_label_dist} and~\Cref{tab:icu_distribution_transposed} detail the label distributions for disease progression identification and general ICU prediction tasks, respectively. The distribution of CXR examinations per patient is presented in~\Cref{tab:stats_cxr_num}.

\begin{table}[h!]\small
    \centering
    \caption{Data statistics in training, validation, and testing sets for each task.}
    \begin{tabular}{l c c c}
        \toprule
          Task & Training & Validation & Test \\
         \midrule
         Disease Progression Identification & 3982 & 560 & 1137 \\
         General ICU Prediction & 1889 & 285 & 546 \\
         \bottomrule
    \end{tabular}
    \label{tab:stats}
\end{table}

\begin{table}[h!]
\small
\setlength{\tabcolsep}{2pt}
\centering
\caption{Label distribution for the Disease Progression Identification task.}
\label{tab:disease_progression_label_dist}
\resizebox{\linewidth}{!}{
\begin{tabular}{l c c c c c c c}
    \toprule
     \makecell[l]{Category} & 
    \makecell{Atelectasis} & 
    \makecell{Enlarged\\Cardiac\\Silhouette} & 
    \makecell{Consolidation} & 
    \makecell{Pulmonary\\Edema} & 
    \makecell{Lung\\Opacity} & 
    \makecell{Pleural\\Effusion} & 
    \makecell{Pneumonia} \\
    \midrule
    Improved    & 328 (15.3\%) & 83 (4.6\%)   & 124 (14.2\%) & 572 (31.9\%) & 784 (19.2\%) & 317 (13.1\%) & 98 (12.7\%) \\
    Worsened    & 674 (31.4\%) & 141 (7.8\%)  & 285 (32.7\%) & 597 (33.3\%) & 1273 (31.2\%) & 702 (28.9\%) & 387 (50.1\%) \\
    No Change   & 1143 (53.3\%) & 1575 (87.5\%) & 463 (53.1\%) & 624 (34.8\%) & 2025 (49.6\%) & 1407 (58.0\%) & 287 (37.2\%) \\
    \midrule
    Total       & 2145 & 1799 & 872 & 1793 & 4082 & 2426 & 772 \\
    \bottomrule
\end{tabular}
}
\end{table}

\begin{table}[h!]
\small
\centering
\caption{Label distribution for the General ICU Prediction Tasks.}
\label{tab:icu_distribution_transposed}
\begin{tabular}{l|cc|cccc}
\toprule
& \multicolumn{2}{c|}{Mortality} & \multicolumn{4}{c}{Length of Stay}\\
\midrule
 & 0 & 1 & $[2, 3)$ & $[3, 4)$ &$[4, 6)$ & $>6$  \\
\midrule
Count    & 2255          & 465           & 793             & 641    &687             & 599              \\
(\%)     & (82.9\%)      & (17.1\%)      & (29.2\%)        & (23.6\%)   &(25.3\%)        & (22.0\%)        \\

\bottomrule
\end{tabular}
\end{table}

\begin{table}[h!]\small
    \centering
    \caption{Data statistics of Numbers of CXR in training, validation, and testing sets for the General ICU Prediction Task.}
    \begin{tabular}{l c c c c}
        \toprule
          Split&2&3&4&5\\
         \midrule
         
Training&1367&465&56&1\\
Validation&199&76&10&0\\
Test&409&120&16&1\\
\midrule
Total&1975&661&82&2\\
         \bottomrule
    \end{tabular}
    \label{tab:stats_cxr_num}
\end{table}

\subsection{Implementation Details}\label{app:implement_details}

\subsubsection{Details of Architectures and Training Procedures of \dipro} \label{app:arch_train}
The training and validation processes are executed on a server equipped with a RTX 3090-24GB GPU card and a 14 vCPU Intel(R) Xeon(R) Gold 6330 CPU. The method is implemented using PyTorch 1.9.1 and PyTorch-Lightning 1.4.2 with CUDA  11.1 environment. AdamW optimizer and CosineAnnealingLR learning rate schedular are used for training. 

\paragraph{Model Architecture and Hyperparameters.}
\dipro consists of four major components: (1) a \textbf{CXR processing module}, which employs a shared ResNet backbone to extract regional visual features, followed by a multi-layer perceptron (MLP) for feature adjustment and two parallel MLP-based projection heads that encode static and dynamic representations; (2) an \textbf{EHR processing module}, which includes a one-layer multivariate transformer encoder for global temporal modeling, a local multi-head attention layer for capturing short-term dependencies, a time-embedding MLP for relative temporal encoding, and a separate MLP for demographic features; (3) a \textbf{multimodal fusion module}, which integrates CXR and EHR representations through local and global attention layers, followed by a lightweight transformer block and a group-based static feature fusion mechanism; and (4) a \textbf{prediction head}, implemented as a task-specific MLP for downstream classification or regression. The detailed hyperparameter settings and implementation specifics are provided in the released code repository.

\paragraph{Training Configuration}
The model was trained with base batch sizes of 8 (for disease progression identification) and 4 (for general ICU prediction), using 4-step gradient accumulation to achieve an effective batch size of 32 or 16. Training proceeded for a maximum of 100 epochs with early stopping triggered after 10 epochs without validation improvement. Task-specific selection metrics were employed: macro-F1 for disease identification, accuracy for length-of-stay classification, and AUPRC for mortality prediction. The hyperparameter search spaces for each task are documented in~\Cref{tab:hyperparams}.

\begin{table}[h]\small
\centering
\caption{Hyperparameter search space used for model tuning.}
\label{tab:hyperparams}
\vspace{2mm}
\begin{tabular}{ll}
\toprule
Hyperparameter& Search Grid \\
\midrule
Learning rate & $8\times10^{-6}$, $5\times10^{-6}$, $1\times10^{-5}$, $5\times10^{-5}$\\
Dropout rate & 0.1, 0.2, 0.3 \\
Hidden dimension & 64, 128, 256 \\
$\lambda_{\text{temp}}$ & 0.01, 0.001, 0.1, 1.0 \\
$\lambda_{\text{pred}}$ & 2, 6, 10 \\
$\lambda_{\text{PAE}}$ &  0.01, 0.1, 2, \\
$\lambda_{\text{orth}}$ &  0.001, 0.01,  0.1, 10 \\
\bottomrule
\end{tabular}
\end{table}

\label{app:implement_details}
\subsubsection{Implementation of Baselines}
\label{app:baselines}
Since none of the baselines can handle both disease identification and general ICU prediction tasks within a unified framework like ours, we introduce minimal modifications to adapt existing approaches. For sequential CXR disease progression specialists (e.g., CheXRelNet~\cite{karwande2022chexrelnet}, CheXRelFormer~\cite{mbakwe2023hierarchical}, and SDPL~\cite{zhu2024symptom}), we concatenate EHR time series and demographic data, then integrate the same EHR encoder and attention fusion layer as our model for fair comparison. For multimodal fusion baselines, we adapt UTDE~\cite{zhang2023improving}, originally designed for longitudinal clinical notes and EHR, by replacing its text encoder with our image encoder to process longitudinal CXRs. Similarly, we modify MedFuse~\cite{hayat2022medfuse} by concatenating CXR representations at the sequence level before fusion, and extend DrFuse~\cite{yao2024drfuse} by first disentangling CXR-EHR pairs and then concatenating the CXR features for fusion.

\paragraph{Hyperparameter Search.}
We conducted a unified hyperparameter search for all baseline models using the following grid: learning rates $\{1\times10^{-5},\,4\times10^{-7},\,1\times10^{-6},\,1\times10^{-7}\}$, dropout rates $\{0.1,\,0.2,\,0.3\}$, and hidden dimensions $\{64,\,128,\,256,\,320\}$. All other hyperparameters were kept consistent with those specified in the original implementations provided by the official source code of each baseline.


\paragraph{Computational Efficiency and Inference Cost Comparison}

We summarize the comparison of computational efficiency and inference cost between \dipro and baseline models in~\Cref{tab:efficiency}, evaluated under the multimodal input setting (sequential CXR + EHR) for the disease progression identification task. Compared to the CheXRelNet baseline, which is also an anatomical region-based model, \dipro reduces FLOPs by 9.8\% and latency by 22.0\%, while achieving a relative gain of 16.5\% in F1 for disease progression.

\begin{table}[h]\small
\centering
\caption{Computational efficiency and inference cost comparison between \dipro and baseline models. (F1 scores are extracted from~\Cref{tab:dp_results-appendix}.)}
\label{tab:efficiency}
\begin{tabular}{lcccccc}
\toprule
Model & 
\makecell{Params\\(M)} & 
\makecell{FLOPs\\(G)} & 
\makecell{MACs\\(G)} & 
\makecell{Latency\\Mean (ms)} & 
\makecell{Throughput\\(samples/s)} & 
\makecell{F1\\Score} \\
\midrule
\dipro (Ours) & 31.06 & 82.75 & 41.37 & 23.05 & 43.38 & \textbf{0.466$\pm$0.018} \\
CheXRelNet~\cite{karwande2022chexrelnet} & 49.15 & 90.90 & 45.45 & 27.34 & 36.58 & 0.382$\pm$0.016 \\
CheXRelFormer~\cite{mbakwe2023hierarchical} & 49.13 & 19.95 & 9.98 & 14.86 & 67.28 & 0.352$\pm$0.021 \\
SDPL~\cite{zhu2024symptom} & 34.24 & 9.45 & 4.73 & 13.35 & 74.90 & 0.393$\pm$0.010 \\
UTDE~\cite{zhang2023improving} & 6.69 & 1.61 & 0.80 & 5.24 & 190.90 &0.449$\pm$0.005\\
UMSE~\cite{lee2023learning} & 23.99 & 8.27 & 4.14 & 9.65 & 103.62 &  0.352$\pm$0.013 \\
MedFuse~\cite{hayat2022medfuse} & 27.21 & 8.28 & 4.14 & 9.78 & 102.23 & 0.409$\pm$0.042 \\
DrFuse~\cite{yao2024drfuse} & 56.40 & 16.45 & 8.23 & 19.93 & 50.18 & 0.429$\pm$0.010 \\
\bottomrule
\end{tabular}
\end{table}

\section{Additional Results}

\begin{table}[ht!]\small
    \centering
    \caption{\textbf{Detailed Performance Comparison on Disease Progression Identification Tasks.} 
Sequential CXR disease progression specialists (e.g., CheXRelNet~\cite{karwande2022chexrelnet}, CheXRelFormer~\cite{mbakwe2023hierarchical}, and SDPL~\cite{zhu2024symptom}), originally designed for unimodal sequential CXR inputs, are extended to multimodal integration by incorporating a transformer-based EHR encoder and applying cross-attention fusion. Methods that can naturally process uni-CXR inputs (e.g., UMSE~\cite{lee2023learning} and MedFuse~\cite{hayat2022medfuse}) are included for comparison under the unimodal CXR setting. 
Numbers in \textbf{bold} indicate the best overall performance, while \underline{underlined} values denote the top-performing baseline.}

    \label{tab:dp_results-appendix}
\resizebox{\linewidth}{!}{
\begin{tabular}{lcccccccc}
\toprule
Method & Precision & Recall & F1 & AUPRC & AUROC & ACC \\
\midrule

\multicolumn{7}{c}{\textbf{Unimodal Methods (CXR)}}\\
   
CheXRelNet~\cite{karwande2022chexrelnet}
    & 0.395$\pm$0.015   & 0.392$\pm$0.010     & 0.389$\pm$0.010    & 0.394$\pm$0.010    & 0.574$\pm$0.011    &  0.508$\pm$0.013
 \\

CheXRelFormer~\cite{mbakwe2023hierarchical}
   & 0.389$\pm$0.044     & 0.379$\pm$0.033     & 0.354$\pm$0.032     & 0.372$\pm$0.023     & 0.551$\pm$0.041   & 0.446$\pm$0.057   \\

SDPL~\cite{zhu2024symptom}
     & 0.408$\pm$0.006    & 0.406$\pm$0.020    &0.393$\pm$0.010   & 0.417$\pm$0.032    & 0.609$\pm$0.031    & 0.538$\pm$0.024
\\

UMSE~\cite{lee2023learning} & 0.337$\pm$0.004     & 0.337$\pm$0.008     & 0.329$\pm$0.008     & 0.347$\pm$0.004     & 0.513$\pm$0.006     & 0.476$\pm$0.004  \\ 
MedFuse~\cite{hayat2022medfuse} & \underline{0.439$\pm$0.006}     & \underline{0.440$\pm$0.009}     & \underline{0.433$\pm$0.010}    & \underline{0.453$\pm$0.011}     & \underline{0.643$\pm$0.006}     & \underline{0.543$\pm$0.024}  \\
\dipro(ours)
 & \textbf{0.475$\pm$0.004}     & \textbf{0.452$\pm$0.011}     & \textbf{0.453$\pm$0.009}     & \textbf{0.468$\pm$0.013 }    &\textbf{ 0.651$\pm$0.016  }    & \textbf{0.567$\pm$0.007}  \\

\midrule
\multicolumn{7}{c}{\textbf{Unimodal Methods (EHR)}}\\
Transformer~\cite{vaswani2017attention} & 0.412$\pm$0.041     & 0.358$\pm$0.010     & 0.327$\pm$0.013     & 0.384$\pm$0.011     & 0.569$\pm$0.008     & 0.509$\pm$0.009 \\

\midrule

\multicolumn{7}{c}{\textbf{Multimodal Methods}}\\
CheXRelNet~\cite{karwande2022chexrelnet} & 0.391$\pm$0.020     & 0.387$\pm$0.019     & 0.382$\pm$0.016     & 0.390$\pm$0.019     & 0.572$\pm$0.028     & 0.504$\pm$0.024
\\

CheXRelFormer~\cite{mbakwe2023hierarchical}  & 0.359$\pm$0.019     & 0.362$\pm$0.026     & 0.352$\pm$0.021     & 0.371$\pm$0.016     & 0.544$\pm$0.022     & 0.474$\pm$0.009 \\

SDPL~\cite{zhu2024symptom} & 0.409$\pm$0.009     & 0.398$\pm$0.006     & 0.393$\pm$0.003     & 0.401$\pm$0.008     & 0.582$\pm$0.008     & 0.529$\pm$0.014
 \\
UTDE~\cite{zhang2023improving}
&  \underline{0.481$\pm$0.017}    & \underline{0.462$\pm$0.002}    & \underline{0.449$\pm$0.005}    & \underline{0.472$\pm$0.014 }   & \underline{0.659$\pm$0.011}& 0.527$\pm$0.016\\

UMSE~\cite{lee2023learning} &  0.353$\pm$0.011     & 0.361$\pm$0.009     & 0.352$\pm$0.013     & 0.364$\pm$0.006     & 0.544$\pm$0.004   & 0.484$\pm$0.011 \\

MedFuse~\cite{hayat2022medfuse} & 0.423$\pm$0.049     & 0.413$\pm$0.045     & 0.409$\pm$0.042     & 0.422$\pm$0.040       & 0.609$\pm$0.05
0  & \underline{0.530$\pm$0.030}\\
DrFuse~\cite{yao2024drfuse}
& 0.442$\pm$0.009     & 0.461$\pm$0.007   & 0.429$\pm$0.010     & 0.438$\pm$0.003     & 0.628$\pm$0.002    & 0.475$\pm$0.021  \\
  
 \dipro(ours)& \textbf{0.484$\pm$0.008}     & \textbf{0.471$\pm$0.024}     & \textbf{0.466$\pm$0.018}     & \textbf{0.478$\pm$0.018 }    & \textbf{0.664$\pm$0.013}  & \textbf{0.565$\pm$0.013}   \\
\bottomrule
\end{tabular}}
\end{table}

\begin{table}[h!]
\small
\setlength{\tabcolsep}{3pt}
\centering
\caption{\textbf{Performance Comparison on Disease Progression Identification Tasks across Different Diseases}.
This table reports the F1 performance of various unimodal and multimodal methods across seven disease progression identification tasks.}
\label{tab:dp_decompose}

\begin{tabular}{lccccccc}
\toprule
\makecell[l]{Method} & Atelectasis & \makecell{Enlarged\\Cardiac\\Silhouette} & Consolidation & \makecell{Pulmonary\\Edema} & \makecell{Lung\\Opacity} & \makecell{Pleural\\Effusion} & Pneumonia \\
\midrule
\multicolumn{8}{c}{\textbf{Unimodal Methods (CXR)}}\\
\hdashline
CheXRelNet~\cite{karwande2022chexrelnet} & 0.425 & \textbf{0.340} & 0.381 & 0.408 & 0.398 & 0.366 & 0.407 \\
CheXRelFormer~\cite{mbakwe2023hierarchical} & 0.359 & 0.318 & 0.319 & 0.390 & 0.354 & 0.380 & 0.357 \\
SDPL~\cite{zhu2024symptom} & 0.396 & 0.362 & 0.350 & 0.439 & 0.431 & 0.443 & 0.331 \\
\dipro (Ours) & \textbf{0.436} & 0.338 & \textbf{0.388} & \textbf{0.527} & \textbf{0.523} & \textbf{0.509} & \textbf{0.452} \\
\midrule
\midrule
\multicolumn{8}{c}{\textbf{Multimodal Methods}}\\
\hdashline
UTDE~\cite{zhang2023improving} & 0.445 & 0.338 & \textbf{0.402} & 0.470 & 0.503 & 0.478 & \textbf{0.510} \\
UMSE~\cite{lee2023learning} & 0.352 & 0.313 & 0.343 & 0.346 & 0.384 & 0.354 & 0.368 \\
MedFuse~\cite{hayat2022medfuse} & 0.422 & 0.340 & 0.392 & 0.455 & 0.447 & 0.433 & 0.363 \\
DrFuse~\cite{yao2024drfuse} & 0.434 & 0.310 & 0.354 & 0.499 & 0.464 & 0.447 & 0.498 \\
\dipro (Ours) & \textbf{0.453} & \textbf{0.362} & 0.399 & \textbf{0.530} & \textbf{0.509} & \textbf{0.500} & \textbf{0.510 }\\
\bottomrule
\end{tabular}
\end{table}

\begin{table}\small
    \centering
    \caption{\textbf{Detailed Performance Comparison on General ICU Prediction Tasks.} 
The ``Input Modalities'' section specifies the data sources used for each method. 
``Last'' and ``Long.'' indicate CXR-only input settings, where ``Last'' denotes the use of the last available CXR and ``Long.'' represents the use of longitudinal CXRs. 
The inclusion of EHR data is indicated by the ``EHR'' column. 
``Sequential CXR Disease Progression Specialists'' are extended to multimodal integration by incorporating a transformer-based EHR encoder and applying cross-attention fusion. 
For ``Clinical Multimodal Fusion Specialists'', which were originally designed for single-CXR inputs, we modify MedFuse~\cite{hayat2022medfuse} by concatenating CXR representations at the sequence level before fusion, and extend DrFuse~\cite{yao2024drfuse} by first disentangling CXR–EHR pairs before concatenating CXR features for multimodal fusion.}

    \label{tab:gen-icu-performance-appendix}
    \resizebox{\textwidth}{!}{%
    \begin{tabular}{lccccccccc}
    \toprule
     &  \multicolumn{3}{c}{Input Modalities} & \multicolumn{2}{c}{Mortality} & \multicolumn{4}{c}{Length of Stay} \\\cmidrule(lr){2-4}\cmidrule(lr){5-6} \cmidrule(lr){7-10}
     Method & Last & Long. &  EHR    & AUPRC       & AUROC  & Kappa & ACC & F1  & AUPRC     \\\midrule
     \multicolumn{10}{c}{\textbf{Unimodal Methods (EHR)}}\\
Transformer~\cite{vaswani2017attention} & & & \coloredcheckmark & 0.712$\pm$0.009     & 0.885$\pm$0.003   & 0.226$\pm$0.019  &0.427$\pm$0.014  & 0.360$\pm$0.024     & 0.386$\pm$0.014  \\
\midrule
\midrule
\multicolumn{10}{c}{\textbf{Sequential CXR Disease Progression Specialists}}\\
\multirow{2}{*}{ChexRelNet~\cite{karwande2022chexrelnet}}&  & \coloredcheckmark  &  & 0.291$\pm$0.050     & 0.624$\pm$0.036  & 0.039$\pm$0.020  & 0.291$\pm$0.014  & 0.238$\pm$0.010   & 0.275$\pm$0.004        \\
  &  & \coloredcheckmark  & \coloredcheckmark & 0.697$\pm$0.040     & 0.876$\pm$0.015   & 0.166$\pm$0.034  & 0.380$\pm$0.028 & 0.355$\pm$0.009     & 0.358$\pm$0.015         \\
    \hdashline
\multirow{2}{*}{ChexRelFormer~\cite{mbakwe2023hierarchical}} &  & \coloredcheckmark  &  & 0.218$\pm$0.011     & 0.522$\pm$0.019  & 0.005$\pm$0.032  & 0.267$\pm$0.043 & 0.212$\pm$0.019     & 0.255$\pm$0.017         \\
  &  & \coloredcheckmark  & \coloredcheckmark & 0.522$\pm$0.041     & 0.766$\pm$0.021 & 0.103$\pm$0.014  & 0.333$\pm$0.005 & 0.306$\pm$0.005     & 0.335$\pm$0.007      \\
    \hdashline
\multirow{2}{*}{SDPL~\cite{zhu2024symptom} } &  & \coloredcheckmark  &   & 0.261$\pm$0.006     & 0.608$\pm$0.035   & 0.011$\pm$0.007 & 0.267$\pm$0.010  & 0.154$\pm$0.022     & 0.261$\pm$0.005         \\
  &  & \coloredcheckmark  & \coloredcheckmark & 0.717$\pm$0.024   & 0.878$\pm$0.019  & \underline{0.231$\pm$0.009} & \underline{0.430$\pm$0.009} & \textbf{0.385$\pm$0.011}     & 0.404$\pm$0.012    \\
 \midrule
\midrule
\multicolumn{10}{c}{\textbf{Longitudinal Multimodal Specialists}}\\
UTDE~\cite{zhang2023improving} & \coloredcheckmark & & \coloredcheckmark &  0.717$\pm$0.019     & 0.887$\pm$0.004 &  0.160$\pm$0.016  & 0.381$\pm$0.013 & 0.324$\pm$0.005     & 0.361$\pm$0.012   \\
    &  & \coloredcheckmark  & \coloredcheckmark &0.710$\pm$0.019     & 0.887$\pm$0.012 & 0.195$\pm$0.031  & 0.400$\pm$0.021 & 0.346$\pm$0.039     & 0.365$\pm$0.028
 \\
   \hdashline
\multirow{2}{*}{UMSE~\cite{lee2023learning}}   & \coloredcheckmark & & \coloredcheckmark& \underline{0.722$\pm$0.039}    & \underline{0.896$\pm$0.012}  & 0.217$\pm$0.013   & 0.419$\pm$0.010  & 0.350$\pm$0.026     & 0.356$\pm$0.018 \\
    &  & \coloredcheckmark  & \coloredcheckmark & 0.712$\pm$0.028     & 
     0.891$\pm$0.011 & 0.204$\pm$0.019 & 0.410$\pm$0.013    & 0.342$\pm$0.021     & 0.357$\pm$0.018\\
      \midrule
\midrule
\multicolumn{10}{c}{\textbf{Clinical Multimodal Fusion Specialists}}\\
\multirow{2}{*}{MedFuse~\cite{hayat2022medfuse}}
       & \coloredcheckmark & & \coloredcheckmark& 0.686$\pm$0.018     & 0.869$\pm$0.011 & 0.213$\pm$0.012  & 0.413$\pm$0.004  & 0.362$\pm$0.025     & \textbf{0.412$\pm$0.010 }
 \\
     &  & \coloredcheckmark  & \coloredcheckmark & 0.716$\pm$0.018     & 0.881$\pm$0.005 &0.210$\pm$0.039   & 0.412$\pm$0.027  & 0.350$\pm$0.006     & \underline{0.410$\pm$0.019} 
 \\
   \hdashline
\multirow{2}{*}{DrFuse~\cite{yao2024drfuse}}
       & \coloredcheckmark & & \coloredcheckmark& 0.709$\pm$0.012     & 0.865$\pm$0.014 & 0.114$\pm$0.048  & 0.338$\pm$0.041 & 0.325$\pm$0.035     & 0.316$\pm$0.024 
 \\
     &  & \coloredcheckmark  & \coloredcheckmark & 0.684$\pm$0.008     & 0.854$\pm$0.017 & 0.142$\pm$0.014 & 0.360$\pm$0.011 & 0.348$\pm$0.006     & 0.329$\pm$0.004 
  \\
       \midrule
\midrule
\multirow{2}{*}{\dipro(Ours)} &  & \coloredcheckmark  &   & 0.319$\pm$0.018    & 0.637$\pm$0.015 & 0.029$\pm$0.017   & 0.284$\pm$0.010      & 0.227$\pm$0.014     & 0.278$\pm$0.004       \\
    &  & \coloredcheckmark  & \coloredcheckmark & \textbf{0.742$\pm$0.003}     & \textbf{0.897$\pm$0.002}   & \textbf{0.248$\pm$0.008} & \textbf{0.440$\pm$0.007} & \underline{0.384$\pm$0.018}     & 0.409$\pm$0.010  \\

    \bottomrule
    \end{tabular}}
\end{table}

\begin{table}[h!]\small
\setlength{\tabcolsep}{2pt}
    \centering
    \caption{Results of the ablation study. (Disease Progression Identification Task)}
    \label{tab:ablation_study_disease_pro_appendix}
    \begin{tabular}{lcccccc}
\toprule
 &  Precision & Recall & F1 & AUPRC & AUROC & ACC \\
 \midrule
\dipro  &  \textbf{0.484$\pm$0.008 }    & \textbf{0.471$\pm$0.024}     & \textbf{0.466$\pm$0.018}     & \textbf{0.478$\pm$0.018}     & \textbf{0.664$\pm$0.013}     & \textbf{0.565$\pm$0.013}\\
\hdashline
w/o MMF & 0.481$\pm$0.009     & 0.460$\pm$0.013     & 0.460$\pm$0.014     & 0.472$\pm$0.008     & 0.654$\pm$0.017     & 0.580$\pm$0.010 \\

w/o PAE & 0.443$\pm$0.024     & 0.446$\pm$0.023     & 0.433$\pm$0.017     & 0.461$\pm$0.015     & 0.646$\pm$0.018     & 0.552$\pm$0.024 \\

w/o STD & 0.372$\pm$0.014     & 0.371$\pm$0.015     & 0.362$\pm$0.016     & 0.377$\pm$0.007     & 0.556$\pm$0.006     & 0.491$\pm$0.027 \\ 

w/o (PAE+MMF) & 0.455$\pm$0.017     & 0.452$\pm$0.015     & 0.439$\pm$0.007     & 0.461$\pm$0.008     & 0.647$\pm$0.003     & 0.536$\pm$0.016 \\

\bottomrule
    \end{tabular}
\end{table}

\begin{table}[h!]\small
\setlength{\tabcolsep}{2pt}
    \centering
    \caption{Results of the ablation study. (Length of Stay Classification)}
    \label{tab:ablation_study_los_appendix}
    \resizebox{\linewidth}{!}{
    \begin{tabular}{lccccccc}
\toprule
 & kappa &  Precision & Recall & F1 & AUPRC & AUROC & ACC \\
 \midrule
 \dipro  & \textbf{0.248$\pm$0.008}     & \textbf{0.431$\pm$0.037}     & \textbf{0.433$\pm$0.005}     & \underline{0.384$\pm$0.018}     & \textbf{0.409$\pm$0.010}     & \textbf{0.688$\pm$0.005}     & \textbf{0.440$\pm$0.007} \\
 \hdashline 
w/o MMF & 0.214$\pm$0.038 & 0.396$\pm$0.053     & 0.408$\pm$0.031     & 0.362$\pm$0.015     & 0.388$\pm$0.009     & 0.673$\pm$0.008     & 0.416$\pm$0.027 \\
 w/o PAE & 0.235$\pm$0.022 & 0.405$\pm$0.021     & 0.423$\pm$0.016     & \textbf{0.386$\pm$0.007}     & 0.408$\pm$0.018     & \textbf{0.688$\pm$0.005 }    & 0.432$\pm$0.018 \\
w/o STD & 0.225$\pm$0.043 & 0.375$\pm$0.082     & 0.415$\pm$0.031     & 0.372$\pm$0.044     & 0.380$\pm$0.025     & 0.667$\pm$0.025     & 0.425$\pm$0.031 \\
w/o (PAE+MMF)&  0.194$\pm$0.014 & 0.367$\pm$0.013     & 0.390$\pm$0.010     & 0.357$\pm$0.009     & 0.380$\pm$0.009     & 0.663$\pm$0.006     & 0.404$\pm$0.014 \\

\bottomrule
    \end{tabular}
    }
\end{table}

\paragraph{Zero-Weight Ablation Study for Loss Penalties}

We conducted ablation experiments in which each loss component, $\lambda_{orth}$ (orthogonal disentanglement loss), $\lambda_{temp}$ (temporal consistency loss for static features), and $\lambda_{PAE}$ (Progression-Aware Enhancement loss), was individually disabled by setting its corresponding weight to zero. The results are summarized in Table~\ref{tab:loss_ablation}. Key observations include:

\begin{itemize}
\item \textbf{Orthogonal disentanglement loss ($\lambda_{orth}$):} Removal caused notable decreases in disease progression and length-of-stay predictions, emphasizing the benefit of disentangling dynamic pathology from static anatomy.
    \item \textbf{Progression-Aware Enhancement loss ($\lambda_{PAE}$):} Disabling this loss reduced disease progression prediction performance, indicating its role in enhancing sensitivity to progression direction.
    \item \textbf{Temporal consistency loss ($\lambda_{temp}$):} Its removal led to significant performance drops across ICU tasks, highlighting the importance of maintaining longitudinal consistency of static anatomical information.
    \item These results demonstrate that all three loss components contribute meaningfully and in distinct ways to the model’s overall performance across different tasks. We report the final selected penalty weights for each prediction task in Table~\ref{tab:penalty_weights}, which reflect the relative contribution of different regularization terms to overall model performance. 
\end{itemize}

\begin{table}[h!]\small
\centering
\caption{Ablation study of loss components by setting individual loss weights to zero. '-' indicates that disease progression prediction could not be computed due to $\mathcal{L}_{\text{temp}}$ requiring at least three CXRs.}
\label{tab:loss_ablation}
\begin{tabular}{lccccccc}
\toprule
&\multicolumn{3}{c}{Components} & Disease Progression & Mortality&Length of stay\\
 \cmidrule(lr){2-4} \cmidrule(lr){5-5} \cmidrule(lr){6-6} \cmidrule(lr){7-7}
ID  & $\lambda_{orth}$  & $\lambda_{PAE}$ &  $\lambda_{temp}$  & F1 &AUPRC  & ACC  \\
\midrule
\dipro & \coloredcheckmark & \coloredcheckmark &  \coloredcheckmark & 0.466 $\pm$ 0.018 & 0.742 $\pm$ 0.003 & 0.440 $\pm$ 0.007 \\
B1    & \xmark  & \coloredcheckmark & \coloredcheckmark  & 0.448 $\pm$ 0.004 & 0.749 $\pm$ 0.028 & 0.415 $\pm$ 0.008 \\
B2    & \coloredcheckmark & \xmark  & \coloredcheckmark & 0.441 $\pm$ 0.013 & 0.708 $\pm$ 0.038 & 0.432 $\pm$ 0.005 \\
B3    & \coloredcheckmark & \coloredcheckmark & \xmark  & -                 & 0.737 $\pm$ 0.008 & 0.415 $\pm$ 0.025 \\
\bottomrule
\end{tabular}
\end{table}
\begin{table}[h!]\small
\centering
\caption{Final selected penalty weights for each prediction task.}
\label{tab:penalty_weights}
\begin{tabular}{lcccc}
\toprule
Prediction Task& $\lambda_{pred}$ & $\lambda_{orth}$ & $\lambda_{temp}$ & $\lambda_{PAE}$ \\
\midrule
In-hospital Mortality & 6 & 0.1 & 1 & 0.1 \\
Length of Stay & 10 & 0.001 & 0.1 & 0.1 \\
Disease Progression & 2 & 1 & -- & 2 \\
\bottomrule
\end{tabular}

\end{table}

\paragraph{Broader Baseline Comparison with Large Vision-Language Models.}To provide a comprehensive evaluation, we include recent large-scale vision-language models (VLMs) capable of processing multi-image inputs (e.g., Gemma3~\cite{team2025gemma}, QWen2.5VL~\cite{bai2023qwen}, and Med-Flamingo~\cite{moor2023med}) as baselines for disease progression identification. We input consecutive CXR pairs and use a few-shot prompting approach (following~\cite{moor2023med}) to have the VLMs predict progression status for thoracic conditions (results in Table~\ref{tab:vlm_comparison}). \dipro achieves notably higher accuracy with lower computational cost, highlighting the advantages of our design: spatialtemporal disentanglement (STD) and progression-aware enhancement (PAE) for clinical prediction.

\begin{table}[h]\small
\centering
\caption{Performance Comparison on Disease Progression Identification Task between VLMs and \dipro.}
\label{tab:vlm_comparison}
\begin{tabular}{lcccc}
\toprule
Model & Precision (P) & Recall (R) & F1 Score & Throughput (samples/s) \\
\midrule
\dipro (Uni-CXR) (Ours) & \textbf{0.475} & \textbf{0.452} & \textbf{0.453} & \textbf{43.38} \\
Gemma3~\cite{team2025gemma} & 0.329 & 0.328 & 0.279 & 0.088 \\
QWen2.5VL~\cite{bai2023qwen} & 0.304 & 0.330 & 0.251 & 0.097 \\
Med-Flamingo~\cite{moor2023med} & 0.355 & 0.345 & 0.301 & 0.234 \\
\bottomrule
\end{tabular}
\end{table}

\paragraph{Robustness to Missing EHR Data} We conducted additional experiments on the general ICU tasks, where EHR data were randomly dropped during training at rates of 25\%, 50\%, and 75\% (following~\cite{wang2023multi}), while testing was performed on the complete dataset. Notably, although model performance naturally decreases as the EHR missing rate increases, \dipro consistently outperforms all established baselines across all missing-data scenarios, demonstrating strong robustness in handling incomplete multimodal inputs (see Tables~\ref{tab:missing_los} and~\ref{tab:missing_mortality}).

\begin{table}[h]\small
\centering
\caption{Length of Stay Classification (Accuracy) under different EHR missing rates.}
\label{tab:missing_los}
\begin{tabular}{lcccc}
\toprule
Method & 75\% Missing & 50\% Missing & 25\% Missing & 0\% Missing \\
\midrule
\dipro (Ours) & \textbf{0.399 $\pm$ 0.011} & \textbf{0.391 $\pm$ 0.021} & \textbf{0.415 $\pm$ 0.005} & \textbf{0.440 $\pm$ 0.007} \\
UMSE~\cite{lee2023learning} & 0.339 $\pm$ 0.029 & 0.376 $\pm$ 0.005 & 0.396 $\pm$ 0.009 & 0.410 $\pm$ 0.013 \\
UTDE~\cite{zhang2023improving}  & 0.377 $\pm$ 0.015 & 0.386 $\pm$ 0.012 & 0.384 $\pm$ 0.034 & 0.400 $\pm$ 0.021 \\
DrFuse~\cite{yao2024drfuse} & 0.340 $\pm$ 0.009 & 0.360 $\pm$ 0.010 & 0.363 $\pm$ 0.011 & 0.360 $\pm$ 0.011 \\
MedFuse~\cite{hayat2022medfuse}& 0.332 $\pm$ 0.039 & 0.360 $\pm$ 0.033 & 0.360 $\pm$ 0.025 & 0.412 $\pm$ 0.027 \\
\bottomrule
\end{tabular}
\end{table}

\begin{table}[h!]\small
\centering
\caption{Mortality Prediction (AUPRC) under different EHR missing rates.}
\label{tab:missing_mortality}
\begin{tabular}{lcccc}
\toprule
Method & 75\% Missing & 50\% Missing & 25\% Missing & 0\% Missing \\
\midrule
\dipro (Ours) & \textbf{0.696 $\pm$ 0.011} & \textbf{0.718 $\pm$ 0.012} & \textbf{0.751 $\pm$ 0.006} & \textbf{0.742 $\pm$ 0.003} \\
UMSE~\cite{lee2023learning} & 0.648 $\pm$ 0.009 & 0.685 $\pm$ 0.010 & 0.686 $\pm$ 0.051 & 0.712 $\pm$ 0.028 \\
UTDE~\cite{zhang2023improving}  & 0.603 $\pm$ 0.052 & 0.673 $\pm$ 0.020 & 0.697 $\pm$ 0.021 & 0.710 $\pm$ 0.019 \\
DrFuse~\cite{yao2024drfuse} & 0.602 $\pm$ 0.024 & 0.663 $\pm$ 0.024 & 0.674 $\pm$ 0.077 & 0.684 $\pm$ 0.008 \\
MedFuse~\cite{hayat2022medfuse} & 0.609 $\pm$ 0.057 & 0.641 $\pm$ 0.026 & 0.705 $\pm$ 0.016 & 0.716 $\pm$ 0.018 \\
\bottomrule
\end{tabular}
\end{table}

\end{document}